\documentclass[letterpaper, 10 pt, conference]{ieeeconf}  %

\IEEEoverridecommandlockouts                              %

\usepackage{graphics} %
\usepackage{epsfig} %
\usepackage{mathptmx} %
\usepackage{times} %
\usepackage{amsmath} %
\usepackage{amssymb}  %
\usepackage{hyperref}
\usepackage{booktabs}
\usepackage[dvipsnames]{xcolor}
\usepackage{lipsum}
\usepackage{xspace}
\usepackage{xcolor}
\usepackage{graphicx}
\usepackage{caption}
\usepackage{subcaption}
\usepackage{adjustbox}
\usepackage{cleveref}
\usepackage{cuted}
\usepackage{arydshln}
\usepackage{circledsteps}
\usepackage[table]{xcolor} 
\usepackage{colortbl}
\definecolor{best}{HTML}{C6D2FF}   %
\definecolor{second}{HTML}{FCE0C5} %

\DeclareMathOperator*{\median}{median}

\newcommand{\ours}{MASt3R-Nav\xspace}
\newcommand{\master}{MASt3R\xspace}
\newcommand{\objr}{ObjectReact\xspace}
\newcommand{\objreact}{ObjectReact\xspace}

\newcommand{\ie}{\emph{i.e.}\xspace}

\title{\LARGE \bf
MASt3R-Nav: WayPixel Navigation in Relative 3D Maps\\
{\footnotesize
Project Page: \url{https://mast3r-nav.github.io/}
}

\author{
Vansh Garg$^{1 \dagger}$,
Rohit Jayanti$^{1 *}$,
Krish Pandya$^{1 *}$,
Sarthak Chittawar$^{1 *}$\\[0.8ex]
Siddharth Tourani$^{2, 3}$,
Muhammad Haris Khan$^3$,
Sourav Garg$^{1 \ddagger}$, and
Madhava Krishna$^{1 \ddagger}$%
  \thanks{The authors acknowledge the support provided by MeitY,
Govt. of India, under the project "Capacity building for
human resource development in Unmanned Aircraft System
(Drone and related Technology)”. This work was also supported in part by Ati Motors.}
\thanks{${\dagger}$ Corresponding Author. $^{*}$ Equal Contribution. $^{\ddagger}$ Equal Advising.}%
  \thanks{
    $^{1}$ Robotics Research Center, IIIT-Hyderabad, India
  }%
\thanks{$^{2}$ University of Heidelberg $^{3}$ MBZUAI}%
}
}

\begin{document}

\maketitle
\thispagestyle{empty}
\pagestyle{empty}

\begin{strip}
    \vspace*{-2cm} %
    \centering
    \includegraphics[width=\textwidth]{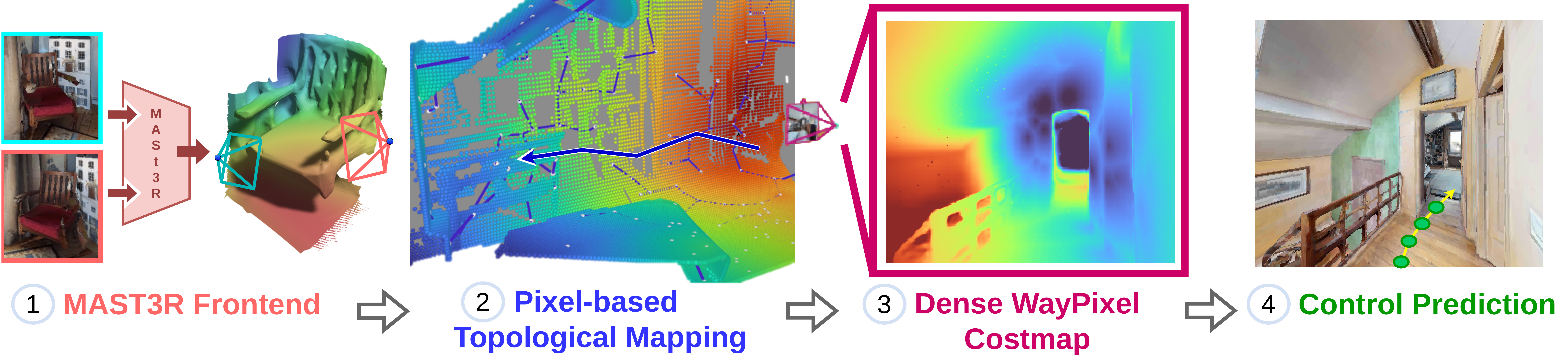}
\captionof{figure}{\textbf{Overview of MASt3R-Nav.}
\Circled{1}~Consecutive RGB frames are matched using MASt3R to obtain dense pixel correspondences and relative 3D point maps.
\Circled{2}~Matched pixels are composed into a pixel-level topological graph, where inter-image correspondences form zero-cost edges, and intra-image edges are weighted by inter-pixel 3D Euclidean distance.
\Circled{3}~Shortest-path planning over this graph yields dense pixel-wise costs for a query view, forming a geometrically informed \textit{WayPixel costmap} with fine-grained gradients toward the goal.
\Circled{4}~A learnt controller conditioned on this dense costmap predicts a trajectory rollout that guides the robot toward the target.}
    \label{fig:teaser}
\end{strip}

\begin{abstract}
Visual navigation ability is strongly tied to its underlying representation of the world. Unlike classical 3D maps that require globally-consistent geometry, image- or object-relative topological graphs almost entirely do away with geometric understanding. But, this comes at the cost of navigation capability, often limiting it to merely teach-and-repeat. In this work, we propose a novel map representation in the form of \textit{pixel-relative connectivity}, which is geometrically accurate but does not require global geometric consistency. Inspired by recent progress in 3D grounded image matching, we construct a map from an image sequence through inter-image connectivity based on pixel correspondences in the relative 3D coordinate systems of individual image pairs. We then use this pixel-level graph to perform global path planning by approximating and sparsifying intra-image pixel connectivity. Through this, we derive a ``\textit{WayPixel Costmap}'' representation and train a controller conditioned on it to predict a trajectory rollout. We show that this dense pixel-level costmap based on relative geometry is a more accurate conditioning variable for control prediction than its image- and object-level counterparts. This enables a highly capable navigation system, as validated on four types of navigation tasks in the simulator and through real world demonstrations. 

\end{abstract}

\section{INTRODUCTION}\label{sec:intro}

Visual navigation requires a structured representation of the surrounding environment that can support direct reasoning for core tasks such as localization, path planning, and control. Classical approaches rely heavily on geometric occupancy maps~\cite{elfes1989using}, where the world is represented in terms of free and occupied space. While such maps enable globally-consistent planning, they typically operate in a metric space and require highly-accurate global registration of 3D points and robot's poses. More recent alternatives propose image-relative topological graphs~\cite{savinov2018semi, shah2023gnm}, where nodes correspond to observed images, and edges encode traversability between them. Most of these metric and image-topological approaches plan paths in the form of 3D waypoints~\cite{huang2023visual} and image goals~\cite{shah2023gnm} respectively. However, these waypoints decouple planning from control: the controller often only has access to a single specific subgoal in either geometric or image space, and that too without any notion of the quality of that subgoal. 

In contrast, recent work has argued for object-relative representations~\cite{garg2024robohop, garg2025objectreact, podgorski2025tango}, where the world is abstracted into semantically meaningful entities, and the object-level path planning costs are directly communicated to the controller. As a result, the controller can learn to be robust to imperfections in planning. However, this benefit is achieved at the cost of reduced geometric understanding due to the object-level abstraction of the underlying map representation.
In this paper, we bridge this gap by introducing a \textit{geometrically precise yet relative} 3D map representation, formulated as dense \textbf{pixel-relative connectivity}, which enables relaying of \textit{pixel-level} path planning costs to the controller (Figure~\ref{fig:teaser} provides an overview). Unlike conventional map abstractions that operate at the level of objects~\cite{garg2025objectreact} or images~\cite{shah2023gnm}, our representation retains pixel-level geometric structure while remaining anchored in relative rather than globally consistent coordinates. This is made possible by recent advances in 3D-grounded pairwise image matching~\cite{wang2024dust3r}, most notably MASt3R~\cite{leroy2024grounding}, whose predicted depth maps and point maps respectively provide complementary sources of intra-image and inter-image pixel connectivity. Leveraging these predictions, we compose a dense connectivity structure across the relative 3D coordinate systems of consecutive image pairs in a traversal sequence, ultimately constructing a graph where each node corresponds to a pixel.

Within this pixel-level graph, edges encode two types of relationships: (1) inter-image correspondences, which form zero-cost edges that directly connect pixels across frames, and (2) intra-image edges, weighted by the 3D Euclidean distance between pixels derived from predicted depth. This formulation enables the use of classical graph search algorithms such as Dijkstra’s algorithm to perform global path planning directly over dense pixel-level structure. To make this representation usable for downstream control, we transform the resulting pixel-level planning costs into a visual costmap that we term the \textit{WayPixel Costmap}. This costmap serves as a fine-grained representation of navigation affordances, encoding subtle geometric cues (e.g., cost gradients on walls, ceilings, or sloped surfaces) that are typically lost in higher-level object-~\cite{garg2025objectreact} or image-relative~\cite{shah2023gnm} abstractions.

\begin{figure}[t!]
    \centering
    \includegraphics[width=\linewidth]{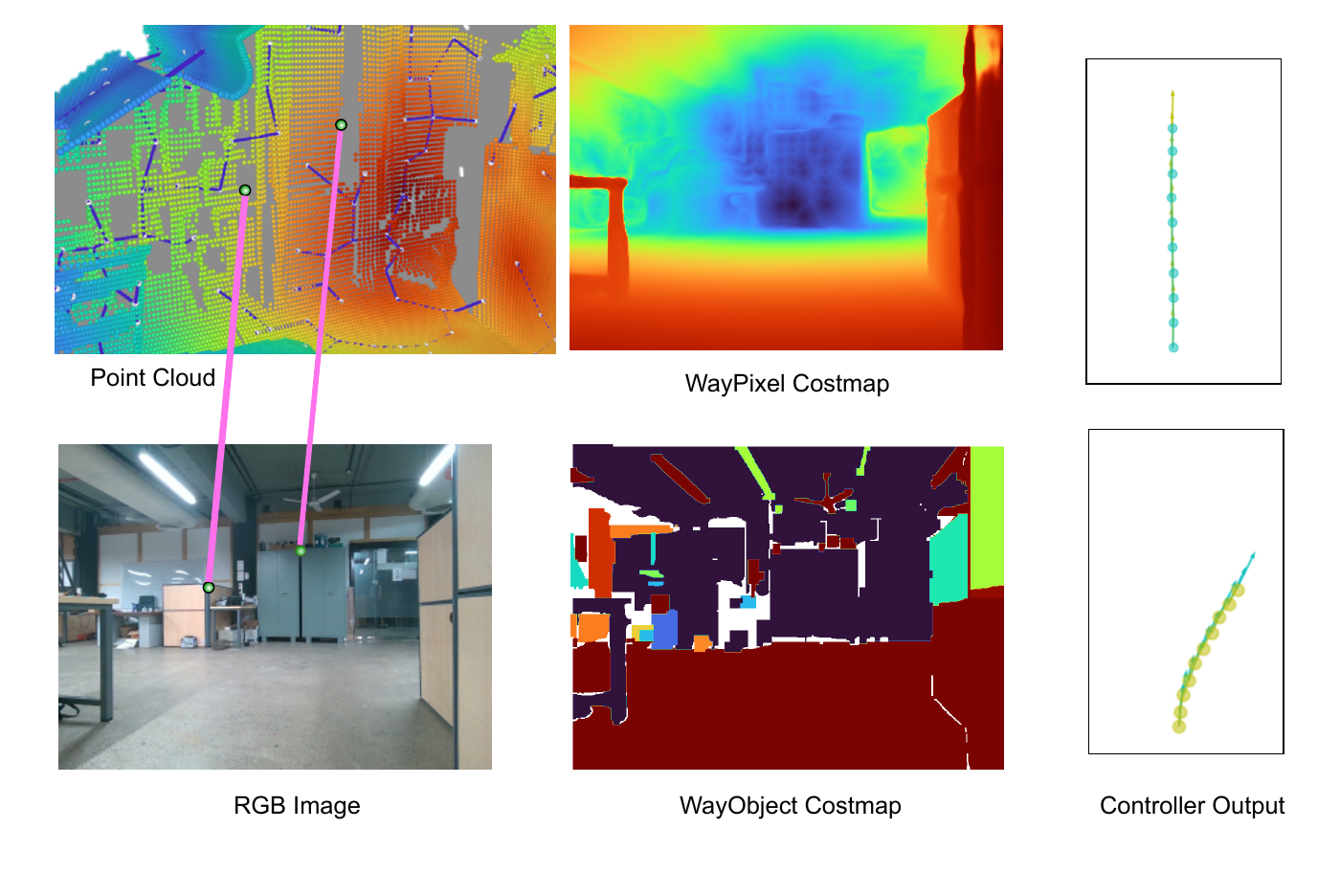}
    \caption{Through our proposed pixel-relative map representa-
tion (top), we obtain dense pixel-level path planning costs for
a given query image (bottom left) in the form of WayPixel
Costmap (top middle). Conditioned on these costmaps, our
proposed PixelReact controller predicts a trajectory rollout
(top right) that is able to correctly guide the robot to the
‘metal cabinet’ goal. On the other hand, the object-relative
representation based coarse WayObject Costmap (bottom
middle) misguides the robot towards the right (bottom right)}
    \label{fig:waypixel_vs_wayobject}
\end{figure}

We then couple this representation with a learning-based controller conditioned on the WayPixel Costmap and predict trajectory rollouts. A key property of this planning-control interface is that the controller can learn to be robust to inconsistencies in the planned path by exploiting local costmap gradients, which is more robust than its object-relative counterpart. This is illustrated in~\Cref{fig:waypixel_vs_wayobject}: we compare pixel-and object-level costmaps, showing how our representation preserves fine-grained geometric structure that correctly leads the robot to the `metal cabinet' goal, unlike the coarseness of object-level abstraction which results in an incorrect right turn.

While pixel-level representations offer fine-grained geometric fidelity, a key limitation of pixel-relative connectivity is its density. A naïve formulation would require a fully connected graph over millions of pixel nodes, which is computationally prohibitive. To make this tractable, we introduce three approximations: (a) intra-image nodes are limited to pixels with at least one inter-image correspondence, ensuring only geometrically grounded pixels are retained; (b) intra-image connectivity is enforced via a Euclidean Minimum Spanning Tree (MST) instead of full connectivity, reducing edge redundancy while preserving distances; and (c) unconnected pixels are selectively re-activated during live operation to bridge localized and connected regions. Together, these approximations reduce graph complexity while maintaining sufficient geometric information for navigation.

With this efficient graph, we construct dense pixel-level costmaps that encode \textit{relative geometry} at high fidelity. These costmaps provide finer gradients and richer local structure than image- or object-level abstractions, yielding a more effective conditioning variable for control. By grounding planning in pixel-relative connectivity, our representation achieves a balance between computational feasibility and geometric precision, enabling more robust and accurate trajectory prediction.

\noindent\textbf{Contributions.} In summary, this paper makes the following key contributions:
\begin{itemize}
    \item We propose \textbf{\master-Nav}, a topological navigation pipeline based on a novel \textbf{pixel-relative representation} that leverages only \emph{local} geometry through relative 3D connectivity of pixels;
    \item We propose a dense \textbf{WayPixel Costmap} representation as an interface between path planning and control, obtained through an efficient method for computing pixel-level path planning costs; and
    \item We propose a new learnt controller, \textbf{PixelReact}, conditioned on the WayPixel Costmaps, enabling it to exploit fine-grained cost gradients for robust trajectory rollout while mitigating the impact of planning errors.
\end{itemize}

We demonstrate that our pixel-relative navigation pipeline can offer both enhanced geometric understanding and improved robustness compared to image- or object-relative navigation paradigms, as validated in the simulator and through real world demonstrations.

\begin{figure*}[t!]
    \centering
    \includegraphics[width=\linewidth]{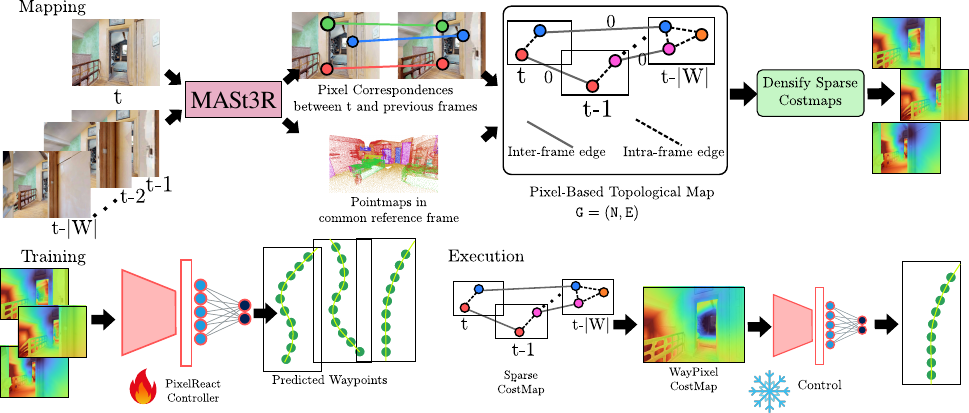}
    \caption{\textbf{MASt3R-Nav Architecture.} \textit{Mapping} phase involves constructing a pixel-level topological graph by linking correspondences across frames and encoding traversal costs using 3D geometry from \master. In the
\textit{Execution} phase, the agent localizes itself against the map and generate a fine-grained pixel costmap by matching current observations and propagating their costs to all pixels.
\textit{Planning} is performed by computing shortest paths through this costmap, yielding dense pixel-wise gradients towards the goal.
Finally, a neural controller consumes the pixel costmap to predict egocentric trajectory waypoints.}
    \label{fig:placeholder}
\end{figure*}

\section{RELATED WORK}

\noindent\textbf{Geometry-Rich 3D Maps:}
A longstanding line of research in visual navigation builds globally-consistent 3D maps that capture accurate geometry. Classical SLAM systems such as ORB-SLAM and LSD-SLAM~\cite{mur2017orb,engel2014lsd,klein2007parallel} reconstruct metric maps, while extensions like SLAM++~\cite{salas2013slam++}, QuadricSLAM~\cite{nicholson2018quadricslam}, and Kimera~\cite{rosinol2021kimera} augment them with semantics. Scene graph formulations~\cite{armeni20193d,rosinol2021kimera} further encode spatial relations between objects and have been used for planning and navigation~\cite{ravichandran2022hierarchical,yin2024sg}. More recently, methods like VGGT~\cite{wang2025vggt}, MASt3R-SLAM~\cite{murai2025mast3r} and MASt3R-Sfm~\cite{duisterhof2025mast3r} have demonstrated 3D reconstruction based on geometry-grounded learning for image matching~\cite{leroy2024grounding,wang2024dust3r}.
While these maps offer geometric precision, they are primarily designed for globally-registered 3D mapping.

\noindent\textbf{Image-Relative Representations:}
To avoid the overhead of global 3D reconstruction, topological methods build maps where nodes correspond to images and edges encode visual or temporal similarity. Inspired by animal landmark navigation, SPTM~\cite{savinov2018semi} proposed such an image graph and trained a controller to reach subgoal \textit{images}. Follow-up works demonstrated extensions to long-horizon tasks~\cite{shah2022viking}, language goals~\cite{shah2022lmnav}, and embodiment generalization~\cite{shah2023gnm}. Local control in this paradigm is obtained by comparing the current view with a chosen subgoal image, either through learning~\cite{li2020learning,meng2020scaling, pathak2018zero}, or visual servoing~\cite{hutchinson2002tutorial,mezouar2002path}. Despite their simplicity, image-relative approaches are tightly coupled to the viewpoints seen during mapping, and errors in subgoal selection can cascade through the controller.  

\noindent\textbf{Object-Relative Representations:}
An alternative line of work grounds navigation in semantically meaningful entities. Recent methods define subgoals at the level of objects, pixels, or points visible in the current observation. RoboHop~\cite{garg2024robohop} and PixNav~\cite{cai2024bridging} respectively use objects and pixels as subgoals, while TANGO~\cite{podgorski2025tango} uses 3D points with occupancy-based traversability estimation. ObjectReact~\cite{garg2025objectreact} uses a \emph{WayObject Costmap} representation based on image segments and their path lengths, and learns a controller conditioned on it. These approaches highlight the benefits of object-relative reasoning: subgoals are invariant to the robot’s pose and can be specified via open-vocabulary queries~\cite{garg2024robohop,huang2023visual}. 
However, their conditioning variables for control remain coarse. RoboHop’s controller is prone to collisions due to simplistic motion priors, PixNav’s learnt policies overfit to layout, and WayObject costmaps discard fine-grained geometric gradients, limiting control precision.  

\noindent\textbf{Pixel-Relative Connectivity:}
Our work takes a different route by moving from image- and object-relative abstractions to a pixel-level representation. Leveraging recent advances in 3D-grounded image matching~\cite{leroy2024grounding}, we construct a graph where nodes are pixels and edges capture relative 3D connectivity both within and across frames. This representation retains geometric fidelity locally without requiring global consistency. From this dense structure, we derive the \emph{WayPixel Costmap}, a fine-grained cost representation that preserves subtle geometry (e.g., wall boundaries, slopes, occlusions) typically lost in image- or object-level abstractions. By conditioning a controller directly on WayPixel costmaps, our approach combines the task relevance of object-level reasoning with the geometric precision of metric maps. In contrast to WayObject costmaps~\cite{garg2025objectreact}, which rely on discrete entities, our pixel-relative formulation enables smoother gradients for control and greater robustness to planning errors.

\section{APPROACH}\label{sec:approach}

Our pixel-relative navigation pipeline is designed to test the hypothesis that a controller conditioned on a fine-grained, pixel-level representation will outperform one conditioned on object- and image-level representations. The framework is structured into three distinct phases: mapping, execution, and training.

\subsection{The \master{} Backbone}
\master~\cite{leroy2024grounding} is a 3D foundation model that takes as input a pair of images and outputs pixel-level correspondences between these images and dense point maps in the form of per-pixel $(x,y,z)$ coordinates for both input images. We utilize both outputs of \master in our pipeline. We denote $\mathtt{D}(p)$ as the point coordinate output by \master corresponding to pixel $p$. We thus denote the 3D distance between the points corresponding to pixels $p$ and $q$ as $\mathtt{D}_{3D}(p,q)=|| \mathtt{D}(p)-\mathtt{D}(q) ||$. Additionally, the important thing to note is that the point maps are output in a shared coordinate frame (that of the first image). Our motivation behind using \master is that relative geometry (as provided by \master) is effective for object-goal navigation and global geometry which can be hard to obtain accurately is not always required. In our pipeline, we use a frozen pre-trained \master throughout the training and evaluation process.

\begin{figure*}[htbp]
    \centering
    \includegraphics[width=1.0\linewidth]{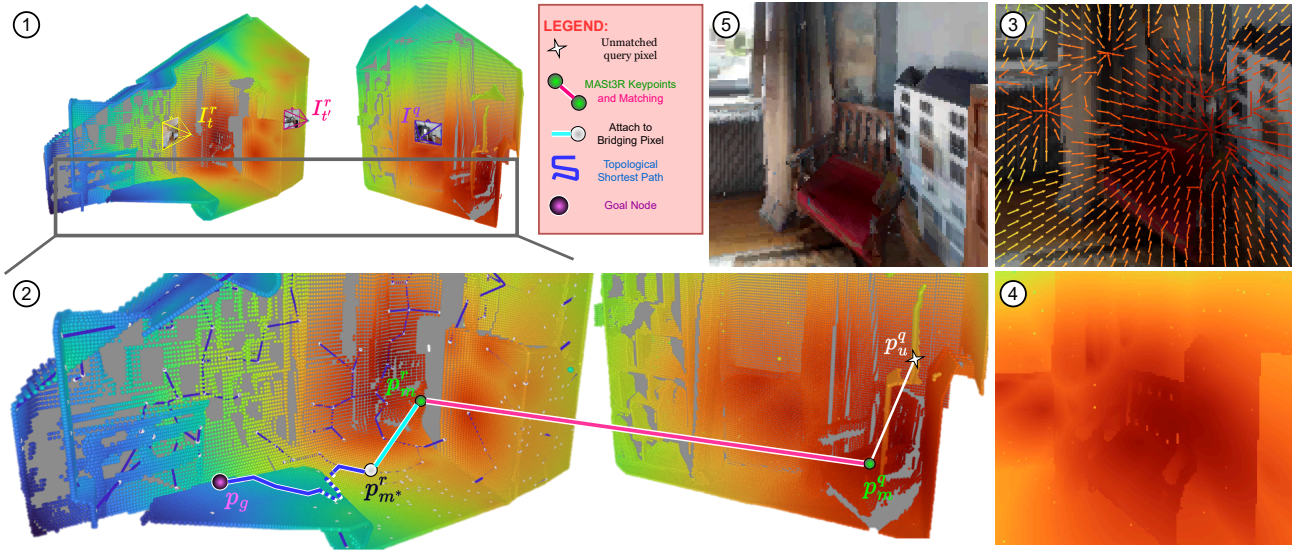}
    \caption{\textbf{WayPixel Costmap generation.} Given the pixel-relative map representation and a query \Circled{1}, we obtain pixel-level planning costs through a series of steps that form a path highlighted in white background \Circled{2} from $p_u^q$ to $p_g$ through $p_m^q$, $p_{m}^r$ and $p_{m^\ast}^r$. \Circled{3} We show the flow of cost gradients from each pixel to its closest least-cost matched pixel and \Circled{4} the final dense WayPixel Costmap on which we condition our trained controller \textit{PixelReact}. \Circled{5} shows the query RGB; the goal position is off-screen, toward the left of the scene.}
    \label{fig:mnav_planning}
\end{figure*}

\subsection{Mapping Phase: Pixel-based Topological Map}
\label{ssec:mapping}

Given a set of reference images from a prior traversal of the environment, we construct a pixel-level topological map. Unlike prior approaches that rely on object segmentation~\cite{garg2025objectreact, garg2024robohop}, our pipeline leverages \master to establish correspondences between consecutive frames. 
Specifically, for the current frame $I_t$, we compute correspondences with a window  $W=\{t-1,t-2,\hdots,t-|W|\}$ of previous frames. The resulting sets of pairwise pixel correspondences are then used to construct a pixel-based graph, denoted as the \textit{Pixel-Based Topological Map}, formally defined as
$\mathtt{G} = (\mathtt{N}, \mathtt{E})$. $\mathtt{N}$ is the set of pixels that participate in correspondences, \ie each matched pixel becomes a node $n \in \mathtt{N}$. 
The edge set $\mathtt{E}$ consists of two categories:

\begin{enumerate}
    \item \textbf{Inter-frame edges:} $e = (p,q) \in \mathtt{E}$ connect a pixel $p$ in the current frame $\mathtt{I}_t$ to its correspondence $q$ in one of the previous frames.  
   These edges are assigned cost $0$~\cite{garg2024robohop}, as they denote the same 3D point observed across frames. Optionally, to prevent unbounded memory growth over long trajectories, we merge pixel node pairs corresponding to these edges such that the merged node is incident to any edge that was incident to the original two nodes. 
   \item \textbf{Intra-frame edges:} $e = (p_1,p_2) \in \mathtt{E}$ connect pairs of pixels within the same frame that each correspond to pixels in other frames.  The cost of these edges is the 3D Euclidean distance $\mathtt{D}_{3D}(p_1,p_2)$ between the points corresponding to $p_1$ and $p_2$ in the \master point cloud. These edges capture the cost of moving between two distinct 3D points within a frame.
\end{enumerate}

More formally, if $\mathtt{C}(\cdot,\cdot)$ denotes the cost between two pixels, we have

{
\begin{align}
    \mathtt{C}(p,q) &= 0 
        \quad \text{if } (p,q) \text{ are matched, } p \in \mathtt{I}_t,\ q \notin \mathtt{I}_t, \\
    \mathtt{C}(p_1,p_2) &= \mathtt{D}_{3D}(p_1,p_2) 
        \quad \text{if } p_1, p_2 \in \mathtt{I}_k,\ k \in \mathtt{W} \cup \{ t  \}
\end{align}
}

Thus, $\mathtt{G}$ encodes costs between 3D points corresponding to a set of matched pixels.  Different strategies can be adopted to determine the connectivity structure of the intra-frame edges  (for example, fully connected graphs or minimum spanning trees). We evaluate some of these choices in the results section.

\subsection{Execution Phase: Localization and Planning}
\label{ssec:execution}

\noindent\textit{Localization:} During the execution phase, the agent must localize itself and plan a path to the goal. At each timestep, the agent captures an RGB image and uses the same dense matcher to find correspondences between its current query frame $I^q_t$ and a submap $\mathtt{S}$. The submap for the current timestep is defined as a sequence of images centered at the localization index; this index is updated at every timestep as a global pointer to the submap image that yields the maximum number of matches with the query image. For our simulator experiments, we use an oracle retriever instead.

\noindent\textit{Planning:} The core of our planning approach is to define a cost for every pixel in the agent's current view. This is achieved through a multi-step process that leverages the pixel-level graph, dense feature matching, and relative 3D geometry. In the following, we explain the computation of pixel-level path planning costs for a given query image, as also illustrated in~\Cref{fig:mnav_planning}.

\paragraph{Topological Path to the Goal} 
The foundation of our proposed costmap is the path from any pixel node $p_j^r$ in the reference image $r$ of the map graph $\mathtt{G}$ to the goal pixel node $p_g$. This path is based on the node connectivity through both the intra-frame and inter-frame edges, and is calculated as a single-source shortest path in the weighted graph using Dijkstra's algorithm. The cost of this path is denoted as $\mathtt{C}(p_j^r, p_g)$. As the map graph and the goal is known apriori, we precompute these costs from every node to the goal node. An example of such a path is shown at the bottom left of~\Cref{fig:mnav_planning}, depicted as the blue-colored path from $p_{m^\ast}^r$ (white pixel node from map image $I_{t'}^r$) to $p_g$ (purple goal node from map image $I_{t}^r$).

\paragraph{Sparse Costmap via Dense Matching}
As shown in~\Cref{fig:mnav_planning}, a query pixel $p_m^q$ (green node in the right point cloud) can get matched with a submap image pixel $p_m^{r}$ (green node in the left point cloud), which is not a connected node in the graph $\mathtt{G}$. Thus, we first find a \textit{bridging map pixel} $p^r_{m^\ast}$ (white node) in the submap image $r$ that minimizes the sum of the local 3D distance and the node's known topological cost: 

\begin{equation} \label{eq:cost_propagation_ref}
\mathtt{C}(p_{m^\ast}^{r}, p_g) = \min_{p^r_k \in \mathtt{N^{r}}} \left( \mathtt{D_{3D}}(p^r_m, p^r_k) + \mathtt{C}(p^r_k, p_g) \right)
\end{equation}
where $\mathtt{N}^{r}$ is the set of graph nodes belonging to the submap image ${r}$. We then use these costs to select a single best matching submap image $r^\ast$ such that it leads the robot towards the goal. We achieve this by minimizing the median of the pixel-level costs per submap image:

\begin{equation}
    r^\ast = \arg\min_{r \in \mathtt{S}} \; \median_{m^\ast} \; \mathtt{C}(p_{m^\ast}^r, p_g)
    \label{eq:min-cost}
\end{equation}
Given the selected submap image ($r^\ast$) and the costs of its pixels that matched with the query image, we directly transfer the costs between the corresponding pixels, \ie, $\mathtt{C}(p_m^q, p_g) = \mathtt{C}(p_{m^\ast}^{r^\ast}, p_g)$. By computing costs for all the matched query pixels, we obtain a \textit{sparse costmap}, which is then converted into a dense representation as described below.

\paragraph{Dense Costmap via Cost Propagation} 
Similar to the object-level costmap representation of ObjectReact~\cite{garg2025objectreact}, we aim to obtain a \textit{dense pixel-level} costmap as a more informative path planning signal for the learnt controller. We refer to this costmap as \textbf{WayPixel Costmap}, defined as an image array where each pixel represents its cost to the goal. Given the sparse costs of only the matched query pixels ($\mathtt{P}_m^q$), we propagate these costs to all the unmatched query pixels ($\mathtt{P}_u^q$). Leveraging the per-pixel 3D point coordinates provided by MASt3R, we can compute the local 3D Euclidean distance $\mathtt{D}_{3D}(p_u^q, p_m^q)$ between an unmatched pixel and all matched pixels in the current view. The cost for an unmatched pixel is then determined by the same cost function shown in Equation~\ref{eq:cost_propagation_ref}, this time using the matched pixels $\mathtt{P}_m^q$ as the \textit{bridging} source of known costs. 
Concretely, the cost for an unmatched pixel is determined by finding a matched neighbor that minimizes the sum of the local 3D distance and the neighbor's known topological cost to the goal, as illustrated with the white colored path from $p_u^q$ (white star) to $p^q_m$ (green node) in~\Cref{fig:mnav_planning}:
\begin{equation}
\label{eq:cost_propagation}
\mathtt{C}(p_u^q, p_g) = \min_{p^q_m \in \mathtt{P}_m^q} \left( \mathtt{D_{3D}}(p^q_u, p^q_m) + \mathtt{C}(p^q_m, p_g) \right).
\end{equation}
This process yields our dense WayPixel Costmap that provides a continuous gradient towards the goal for every pixel in the agent’s field of view. Similar to ObjectReact~\cite{garg2025objectreact}, to create a representation suitable for a neural network, these scalar costs are then encoded into a higher-dimensional embedding using sinusoidal functions, analogous to positional encodings~\cite{shaw2018self} in Transformers~\cite{vaswani2017attention}.

\subsection{Training Phase: The PixelReact Controller}
\label{ssec:training}

Finally, we propose a WayPixel Costmap conditioned learnt controller, \textbf{PixelReact}, which learns to predict a sequence of future waypoints from the dense pixel-level costs. To ensure a fair comparison and isolate the contribution of the novel representation, we adopt the neural network architecture and imitation learning procedure from the \objr~\cite{garg2025objectreact} baseline. The controller architecture consists of a convolutional encoder that processes the \textit{Pixel Costmap} and a small MLP decoder that outputs local waypoints, that is, position and yaw relative to the current robot position. Specifically, this prediction takes the form of a trajectory rollout consisting of 10 future 2D waypoints in the robot's local bird's eye view (BEV) space. We train the model using the same data as ObjectReact~\cite{garg2025objectreact} based on the trajectories from the Habitat simulator (HM3Dv0.2 dataset). The training objective is a regression loss, specifically the L2 distance, between the predicted waypoints $\hat{\mathbf{y}}$ and the ground-truth shortest-path waypoints $\mathbf{y}$. A rigorous, direct comparison is made against the \objr~controller, trained under identical conditions but with its original object-based costmap as input.

\begin{table}[t]
\centering
\caption{Comparison of object-level vs. pixel-level representations 
for the Imitate task.}
\label{tab:rep_comparison}
\begin{tabular}{l l l c c}
\toprule
Mapper & Localizer & Controller & SPL & SSPL \\
\midrule
\multicolumn{5}{c}{\textbf{Object-level Representation}} \\[0.3em]
LGlue & LGlue     & ObjectReact & 51.51 & 58.59 \\
\master~ & LGlue   & ObjectReact & 45.45 & 53.21 \\
LGlue & \master   & ObjectReact & 51.50 & 60.85 \\
\master~ & \master & ObjectReact & 51.48 & 58.64 \\
\midrule
\multicolumn{5}{c}{\textbf{Pixel-level Representation}} \\[0.3em]
\master~ & \master & ObjectReact & 63.63 & 74.11 \\
\master~ & \master & PixelReact  & \cellcolor{best}\textbf{81.77} & \cellcolor{best}\textbf{84.36} \\
\bottomrule
\end{tabular}
\end{table}

\begin{table*}
\centering
\caption{\textbf{State-of-the-art} comparison of different control methods on four navigation tasks. \colorbox{best}{Blue} cells mark the best scores; \colorbox{second}{Orange} cells mark the second-best.}
\begin{tabular}{l cc cc cc cc cc cc} 
\toprule
  &&& \multicolumn{2}{c}{\textbf{Imitate}} & \multicolumn{2}{c}{\textbf{Alt Goal}} & \multicolumn{2}{c}{\textbf{Shortcut}} & \multicolumn{2}{c}{\textbf{Reverse}} & \multicolumn{2}{c}{\textbf{\textit{Average}}} \\
\cmidrule(lr{0.75em}){4-13}

\textbf{Method} & \textbf{Type} & \textbf{Train Data} &
  SPL & SSPL & SPL & SSPL & SPL & SSPL & SPL & SSPL & SPL & SSPL \\

\cmidrule(lr{0.75em}){1-3}
\cmidrule(lr{0.75em}){4-5}
\cmidrule(lr{0.75em}){6-7}
\cmidrule(lr{0.75em}){8-9}
\cmidrule(lr{0.75em}){10-11}
\cmidrule(lr{0.75em}){12-13}

GNM~\cite{shah2023gnm} & Image-Relative & Real 
& 78.79 & 82.95 & 8.70 & 15.44 & 15.38 & 31.74 & 3.33 & 6.11 
& 26.55 & 34.56 \\

GNM (HM3D)~\cite{garg2025objectreact} & Image-Relative & HM3D 
& \cellcolor{second}81.82 & \cellcolor{second}86.38 & 0.00 & 10.91 & 15.38 & 24.57 & 13.28 & 20.77 
& 27.62 & 35.66 \\

\midrule
PixNav~\cite{cai2024bridging} & Object-Relative & HM3D 
& 42.42 & 46.75 & 26.09 & 31.66 & 7.69 & 22.29 & 16.16 & 25.56 
& 23.09 & 31.57 \\

RoboHop~\cite{garg2024robohop} & Object-Relative & Zero Shot 
& 57.56 & 64.99 & \cellcolor{second}30.43 & \cellcolor{second}38.23 & \cellcolor{second}30.77 & \cellcolor{second}40.87 & 9.98 & 16.92 
& 32.19 & 40.25 \\

ObjectReact~\cite{garg2025objectreact} & Object-Relative & HM3D 
& 60.60 & 68.51 & 21.74 & 26.68 & 23.08 & 39.64 & \cellcolor{best}30.00 & \cellcolor{best}42.01 
& \cellcolor{second}33.36 & \cellcolor{second}44.71 \\

\midrule
\master-Nav (\textbf{Ours}) & Pixel-Relative & HM3D 
& \cellcolor{best}93.94 & \cellcolor{best}94.95 & \cellcolor{best}47.83 & \cellcolor{best}58.06 & \cellcolor{best}46.15 & \cellcolor{best}61.10 & \cellcolor{second}23.25 & \cellcolor{second}26.83 
& \cellcolor{best}52.79 & \cellcolor{best}60.24 \\

\bottomrule
\end{tabular}
\label{tab:sota}
\end{table*}

\newcommand{\ph}{\textbf{--}}

\begin{table*}
\centering
\caption{\textbf{Representation Efficiency} comparing inter- and intra-frame connectivity strategies.}
\begin{tabular}{l l c c c c c c c c}
\toprule
\multicolumn{2}{c}{\textbf{Connectivity}} & \multicolumn{3}{c}{\textbf{Topomap Graph Stats}} & \multicolumn{3}{c}{\textbf{Computation Time (s)}} & \multicolumn{2}{c}{\textbf{Navigation}} \\
\cmidrule(lr{0.5em}){1-2}
\cmidrule(lr{0.5em}){3-5}
\cmidrule(lr{0.5em}){6-8}
\cmidrule(lr{0.5em}){9-10}

\textbf{EC} & \textbf{NC} &
   \textbf{Num Nodes} & \textbf{Intra-Frame Edges} & \textbf{Disk (MB)} &
   \textbf{Intra} & \textbf{Edge-Weight} & \textbf{Dijkstra} &
   \textbf{SPL} & \textbf{SSPL} \\
\midrule
Exhaustive & Sub10 & 24076 & 4660494 & 90.6 & 16.7 & 50.5 & 9.0 & 74.78 & 81.41 \\
EMST & Sub10 & 24076 &   24011 & \textbf{73.9} &  8.7 & \textbf{2.1} & \textbf{1.4} & \textbf{78.62} & \textbf{82.85} \\
Delaunay 3D & Sub10 & 24076 &  167970 & 75.3 &  \textbf{7.0} &  3.7 & 1.7 & 78.56 & \textbf{82.85} \\
EMST & None & 191281 & 191215 & 78.4 & 98.2 & 4.3 & 2.3 & 62.06 & 73.16 \\
Delaunay 3D & None & 191281 & 1418359 & 85.9 & 25.7 & 24.2 & 8.5 & 66.28 & 74.92 \\
\midrule
\multicolumn{2}{l}{ObjectReact~\cite{garg2025objectreact}} & 1676 & 22633 & 4.76 & \ph & \ph & 0.011 & 60.60 & 68.51 \\
\bottomrule
\end{tabular}
\label{tab:scalability}
\end{table*}

\section{EXPERIMENTAL SETUP}
We evaluate navigation performance on the HM3D-IIN dataset, using its validation split~\cite{garg2025objectreact, podgorski2025tango}, which consists of $36$ unique scenes, with one episode per scene. For every episode, a prior map is available in the form of an image sequence.
In the offline mapping phase, we construct a pixel-relative 3D map using these RGB images and pointmaps obtained from \master. During the online execution phase, the agent is initialized along the map trajectory in a way that it starts at least $5m$ (geodesic distance) or further away from the goal, ensuring that the path requires traversing multiple rooms and corridors on the same floor of the house. Throughout all our experiments, the agent is provided with ground-truth localization in the form of the map image index that is closest to its current 2D position.

\noindent\textbf{Evaluation Metrics:}
We measure the controller's ability to reach the object goal within an episode. An episode is considered successful if the agent reaches within $1m$ of the goal in at most $300$ steps, with an oracle stop condition. We report Success weighted by Path Length (SPL)~\cite{anderson2018evaluation} and Soft-SPL (SSPL)~\cite{datta2021integrating} metrics as in \objreact. SPL is a strict metric designed to measure both task completion and navigation efficiency, calculated by weighting a binary success indicator by the ratio of the shortest path distance to the maximum of the shortest and actually traversed path lengths. To account for near-misses, we also report SSPL, which replaces the binary success term with a soft value indicating the progress made by the agent toward the goal. This is particularly useful for episodes that are deemed failures (SPL$=0$) but have progressed toward the goal (SSPL$>0$). The results are averaged over $36$ runs, using a fixed sensor height of $0.4m$ for both mapping and execution phases.

\section{RESULTS}

\subsection{Object Relative vs. Pixel Relative}
Table~\ref{tab:rep_comparison} ablates different modules of our navigation pipeline in comparison to ObjectReact~\cite{garg2025objectreact} (first row). It can be observed that enhancing ObjectReact by simply replacing LightGlue (default for ObjectReact) with \master as a matcher for mapping and localization (Rows 2-4) does not lead to any major performance gain. This can be attributed to a poor propagation of \textit{dense} pixel-level matches to object-level matches, which subsequently affects the quality of WayObject Costmap. On the other hand, our pixel-relative map representation generates fine-grained WayPixel Costmaps, which not only predict significantly better control using the original ObjectReact controller (penultimate row) but lead to a very high navigation performance with our PixelReact controller (last row). For these experiments, localization was performed against submaps constructed using a temporal radius of 16 frames with a subsampling factor of 2.

\subsection{State-of-the-art comparisons on multiple tasks}
In Table~\ref{tab:sota}, we compare our proposed method against several existing methods on four challenging tasks as proposed in \objreact~\cite{garg2025objectreact}. In these experiments, the localization submaps are constructed with a temporal radius of 4 frames and a subsampling factor of 1.
\begin{itemize}
    \item \textbf{Imitate} requires the agent to follow it's prior trajectory, similar to a teach-and-repeat setup. 
    \item \textbf{Alt Goal} involves navigating to a previously observed but unvisited goal, requiring the agent to take a new route. 
    \item \textbf{Shortcut} modifies the mapping trajectory by including an additional stop at the Alt Goal, such that agent must find a shorter path to the final goal. 
    \item \textbf{Reverse} tests the agent's ability to traverse the prior trajectory in the opposite direction.
\end{itemize}
Our method outperforms all the baselines on the Imitate, AltGoal and Shortcut tasks by large margins on both the metrics. On the Imitate task, we outperform the previous best method GNM~\cite{shah2023gnm} by absolute $10\%$ on both SPL and SSPL. A similar trend holds against the GNM model trained on HM3D~\cite{garg2025objectreact}, using exactly the same training data as ours. On the AltGoal and Shortcut task, object-relative approaches substantially outperform image-relative baselines. On the other hand, our pixel-relative approach almost doubles the SPL and SSPL achieved by the object-relative methods. On the challenging reverse task, we achieve low performance, which is attributed to \master's incorrect selection of a reference image for path planning. Although \master{} like methods are demonstrated to work under large viewpoint shifts, we observed that it often found similar number of inliers for a true positive and a false positive match under the duress of low visual overlap. Nevertheless, our method achieves superior performance on average across all tasks, demonstrating its effectiveness in general.

\subsection{Representation Efficiency for Pixel-Relative 3D Maps}
Table~\ref{tab:scalability} compares the extent of trade-off between representation efficiency and navigation success. Here, EC and NC denote the edge and node culling strategies respectively, while the reported graph statistics correspond to the number of retained nodes, intra-frame edges, and graph storage size. We additionally report the computation time for intra-frame edge construction, edge-weight computation, and shortest-path planning, along with the resulting navigation performance. We consider three types of intra-frame connections: \texttt{Exhaustive}, \ie, a complete intra-frame graph with edges between all pixel nodes; \texttt{Delaunay 3D}, 
where edges are based on the circumscribing criteria of Delaunay Triangulation; and \texttt{EMST}, \ie, Euclidean Minimum Spanning Tree, with minimum number of edges to maintain a connected graph. It can be observed that the number of edges drop from 4.6 million to merely 24,011 when replacing exhaustive connectivity with EMST. This significantly reduces the planning time while maintaining similar performance trends for navigation success. Since \master is a dense image matching method, we randomly subsample pixel correspondences by a factor of $10$, which proportionally reduces the number of pixel nodes that get created in any given image. In Table~\ref{tab:scalability}, we also consider a baseline that uses all the pixel correspondences (2nd and 3rd last rows with `None' subsampling); we consider both EMST and Delaunay 3D based intra-frame edge connectivity for this baseline. It can be observed that using all the correspondences substantially increases the number of nodes and edges, resulting in a high computation time for planning while reducing navigation success metrics. We finally compare against the ObjectReact baseline for additional context, which is much more efficient but not as performant. Overall, this study highlights that point-level geometry with sparse connectivity suffices for robot navigation, and can potentially scale to even larger environments.

\begin{table}[t]
\centering
\caption{Scalability of runtime and memory with increasing scene size. Graph complexity is denoted as $I$ / $V$ / $E$, representing the number of Images ($I$), Vertices/Nodes ($V$), and Edges ($E$).}
\label{tab:scalability_node_merging}
\vspace{-2mm}
\setlength{\tabcolsep}{3pt}
\renewcommand{\arraystretch}{1.05}
\resizebox{\columnwidth}{!}{%
\begin{tabular}{@{} l c c c @{}}
\toprule
& \multicolumn{3}{c}{\textbf{Scene Scale}} \\
\cmidrule(l){2-4}
\textbf{Metric} 
& \textbf{One Floor} 
& \textbf{Two Floors} 
& \textbf{Three Floors} \\
\midrule
$I$ / $V$ / $E$ (Ours)
& 178\,/\,21{,}810\,/\,42{,}096
& 417\,/\,51{,}823\,/\,100{,}132
& 639\,/\,80{,}217\,/\,154{,}865 \\
$I$ / $V$ / $E$ (ObjectReact)
& 178\,/\,3{,}031\,/\,12{,}127
& 417\,/\,6{,}949\,/\,28{,}162
& 639\,/\,9{,}352\,/\,37{,}503 \\
\midrule
\multicolumn{4}{@{}l}{\textit{Runtime (s)}} \\
\quad Depth Estimation
& 18.29 & 43.21 & 66.89 \\
\quad Matching
& 54.13 & 130.39 & 202.15 \\
\quad Graph Construction
& 0.98 & 2.63 & 4.15 \\
\quad Planning
& 0.04 & 0.11 & 0.18 \\
\quad Dense Costmaps
& 2.22 & 5.41 & 8.68 \\
\cmidrule(l{1em}){1-4}
\quad \textbf{Total (Ours)}
& \textbf{78.74} & \textbf{189.01} & \textbf{294.41} \\
\quad Total (ObjectReact)
& 36.90 & 87.89 & 136.09 \\
\midrule
\multicolumn{4}{@{}l}{\textit{Memory (MB)}} \\
\quad Graph
& 1.20 & 2.93 & 4.53 \\
\quad MASt3R Pointmap
& 149.70 & 350.40 & 535.65 \\
\cmidrule(l{1em}){1-4}
\quad \textbf{Total (Ours)}
& \textbf{150.80} & \textbf{353.33} & \textbf{540.18} \\
\bottomrule
\end{tabular}%
}
\vspace{-4mm}
\end{table}

\begin{figure*}[ht]
    \centering
    \includegraphics[width=1.0\linewidth]{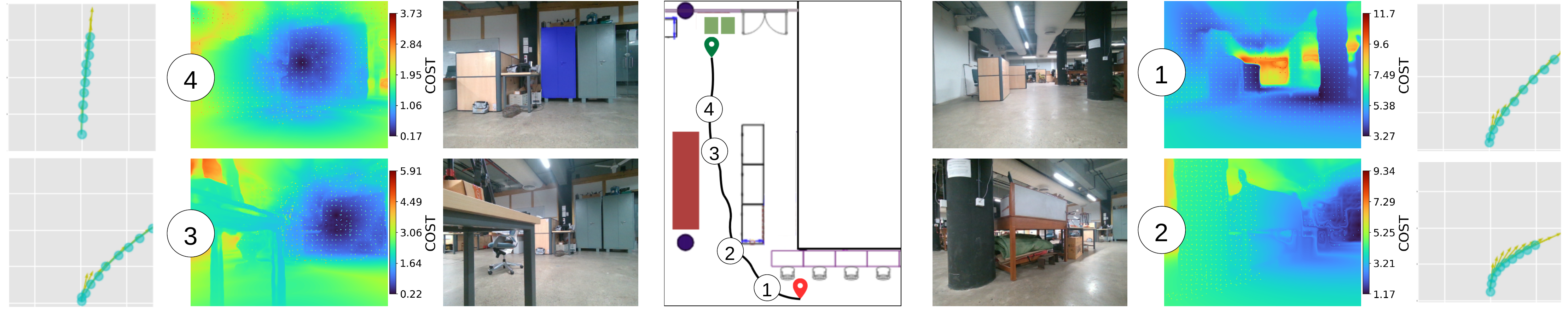}
    \caption{\textbf{Real World Demonstration.} We show RGB observations, their WayPixel costmaps and the controller waypoints towards the goal object (shaded in \textcolor{blue}{blue} in image 4) on four different locations in the robot trajectory.}
    \label{fig:real_world_1}
\end{figure*}

\subsection{Scalability: Computational and Storage Footprint}
We evaluate the computational and storage requirements of our offline mapping and planning pipeline as the environment scale increases, using a video captured across a three-floor, 30×36m office building~\cite{garg2025objectreact}. To ensure a fair comparison, all evaluations, including baseline runtimes for ObjectReact, were conducted on a server with an AMD Ryzen 9 7950X CPU and an NVIDIA RTX A4000 (16GB) GPU. As shown in Table~\ref{tab:scalability_node_merging}, both the total runtime and memory consumption scale roughly linearly with the expanding size (floors) of the environment. With over 80,000 pixel nodes, \master-Nav's map generation and dense costmap computation executes in under five minutes (294.41s), requiring 540.18 MB of storage.

\subsection{Real-World Demonstration}
To validate the practical applicability and sim-to-real transfer capabilities of our proposed method, we deployed our navigation pipeline in real world. We used a P3DX mobile robot, equipped with a RealSense camera for RGB images.
Figure~\ref{fig:real_world_1} shows a topdown view at the center, showcasing the robot's successful trajectory from the start marker (red) to the goal marker (green). We also show the RGB images, their corresponding WayPixel Costmaps, and the trajectory rollout predicted using our \ours{} controller.
It is clear that despite being trained exclusively on the HM3D simulated dataset, our navigation pipeline performs effectively during inference on a real-world mobile robot in an unseen environment.

\section{CONCLUSION}
We presented a visual navigation pipeline, \master-Nav, with several desirable characteristics: a robust perception module driven by advances in 3D grounded image matching; a novel pixel-relative geometric representation for mapping and global path planning; an intermediate WayPixel Costmap representation as dense and informative interface between planning and control; and a learnt controller, PixelReact, conditioned on WayPixel Costmaps to predict a trajectory rollout. Our pixel-relative approach surpasses prior state-of-the-art on multiple challenging navigation tasks, outperforming both image-relative and object-relative methods. This establishes the efficacy of relative pixel-level geometry for navigation without the need for globally-consistent 3D reconstruction, pose estimation, sensor depth or traversability estimation. Future work can consider a hybrid representation approach that combines the benefits of object-level abstraction and pixel-level geometry to solve highly challenging tasks such as reverse path tracing.

\addtolength{\textheight}{-4cm}   %

\bibliographystyle{IEEEtran}
\bibliography{references,reference_sg}

\end{document}